\documentclass[lettersize,journal]{IEEEtran}
\usepackage{amsmath,amsfonts}
\usepackage{algorithmic}
\usepackage{algorithm}
\usepackage{array}
\usepackage[caption=false,font=normalsize,labelfont=sf,textfont=sf]{subfig}
\usepackage{textcomp}
\usepackage{stfloats}
\usepackage{url}
\usepackage{verbatim}
\usepackage{graphicx}
\usepackage{cite}
\usepackage{multirow}
\usepackage{amsmath}
\usepackage{amssymb}
\usepackage{bbding}
\usepackage{amssymb}
\usepackage{amsfonts}
\usepackage{hyperref}

\begin{document}

\title{Training-Free Large Model Priors for Multiple-in-One Image Restoration}

\author{Xuanhua He, Lang Li, Yingying Wang, Hui Zheng, Ke Cao, Keyu Yan,\\
Rui Li, Chengjun Xie, Jie Zhang, Man Zhou
\thanks{     This work was supported by the National Natural Science Foundation of China under grant number 32171888 and HFIPS Director's Fund under grant No.2023YZGH04 . Xuanhua He and Lang Li contributed equally; Corresponding author: Jie Zhang and Man Zhou;

Xuanhua He, Lang Li, Keyu Yan and Ke Cao are with  Hefei Institutes of Physical Science, Chinese Academy of Sciences, Hefei 230031 and also with University of Science and Technology of China, Hefei 230026, (e-mail: hexuanhua, caoke200820, keyu@mail.ustc.edu.cn);

Yingying Wang and Hui Zheng are with Key Laboratory of Multimedia Trusted Perception and Efficient Computing, Xiamen University, Xiamen 361102, China (e-mail: wangyingying7, huiiz@stu.xmu.edu.cn);

Man Zhou is University of Science and Technology of China, China  (e-mail:manman@mail.ustc.edu.cn, manzhountu@gmail.com);

Jie Zhang, Rui Li and Chengjun Xie is with the Intelligent Agriculture Engineering Laboratory of Anhui Province, Institute of Intelligent Machines, and Hefei Institutes of Physical Science, Chinese Academy of Sciences, Hefei 230031, China (e-mail: zhangjie, lirui, cjxie@iim.ac.cn;);
}
}

\markboth{Journal of \LaTeX\ Class Files,~Vol.~14, No.~8, August~2021}%
{Shell \MakeLowercase{\textit{et al.}}: A Sample Article Using IEEEtran.cls for IEEE Journals}

\maketitle

\begin{abstract}
Image restoration aims to reconstruct the latent clear images from their degraded versions. Despite the notable achievement, existing methods predominantly focus on handling specific degradation types and thus require specialized models, impeding real-world applications in dynamic degradation scenarios.  
To address this issue, we propose Large Model Driven Image Restoration framework (LMDIR), a novel multiple-in-one image restoration paradigm that leverages the generic priors from large multi-modal language models (MMLMs) and the pretrained diffusion models. In detail, LMDIR integrates three key prior knowledges: 1) global degradation knowledge from MMLMs, 2) scene-aware contextual descriptions generated by MMLMs, and 3) fine-grained high-quality reference images synthesized by diffusion models guided by MMLM descriptions. 
Standing on above priors, our architecture comprises a query-based prompt encoder, degradation-aware transformer block injecting global degradation knowledge, content-aware transformer block incorporating scene description, and reference-based transformer block incorporating fine-grained image priors. This design facilitates single-stage training paradigm to address various degradations while supporting both automatic and user-guided restoration. Extensive experiments demonstrate that our designed method outperforms state-of-the-art competitors on multiple evaluation benchmarks.
\end{abstract}

\begin{IEEEkeywords}
All-in-one Image Restoration, Large Model, Diffusion Model.
\end{IEEEkeywords}

\section{Introduction}
Image restoration, a classical low-level vision task, aims to reconstruct the latent high-quality images from their corrupted counterparts affected by various types of degradation, such as rain streaks~\cite{rain1,rain2}, low-light conditions~\cite{zerodce,ren2019low}, and noise~\cite{li2023ntire,chen2023masked}. Traditional image restoration methods have developed various natural image priors, \emph{e.g.,} low-rank prior and total variation regularization \cite{7473901, 2009Variational} to regularize the solution space of the latent clear image. However, designing and optimizing these priors is challenging, which limits their practical applicability. The advent of deep learning has brought significant advancements in the field of image restoration. However, in real-world scenarios, such as autonomous driving or surveillance monitoring, degradation types can be random and time-varying, resulting in a wide range of distortions across different scenes.

Existing image restoration models are predominantly tailored to handle specific degradation types, necessitating the training of specialized models for each type of degradation~\cite{liang2021swinir,zamir2022restormer}. Furthermore, these methods require complex mechanisms to match the input degraded image with the appropriate restoration model. This paradigm impedes the application of image restoration techniques in real-world applications, where the degradation type is often dynamic.
\begin{figure*}
    \centering
    \includegraphics[width=\linewidth]{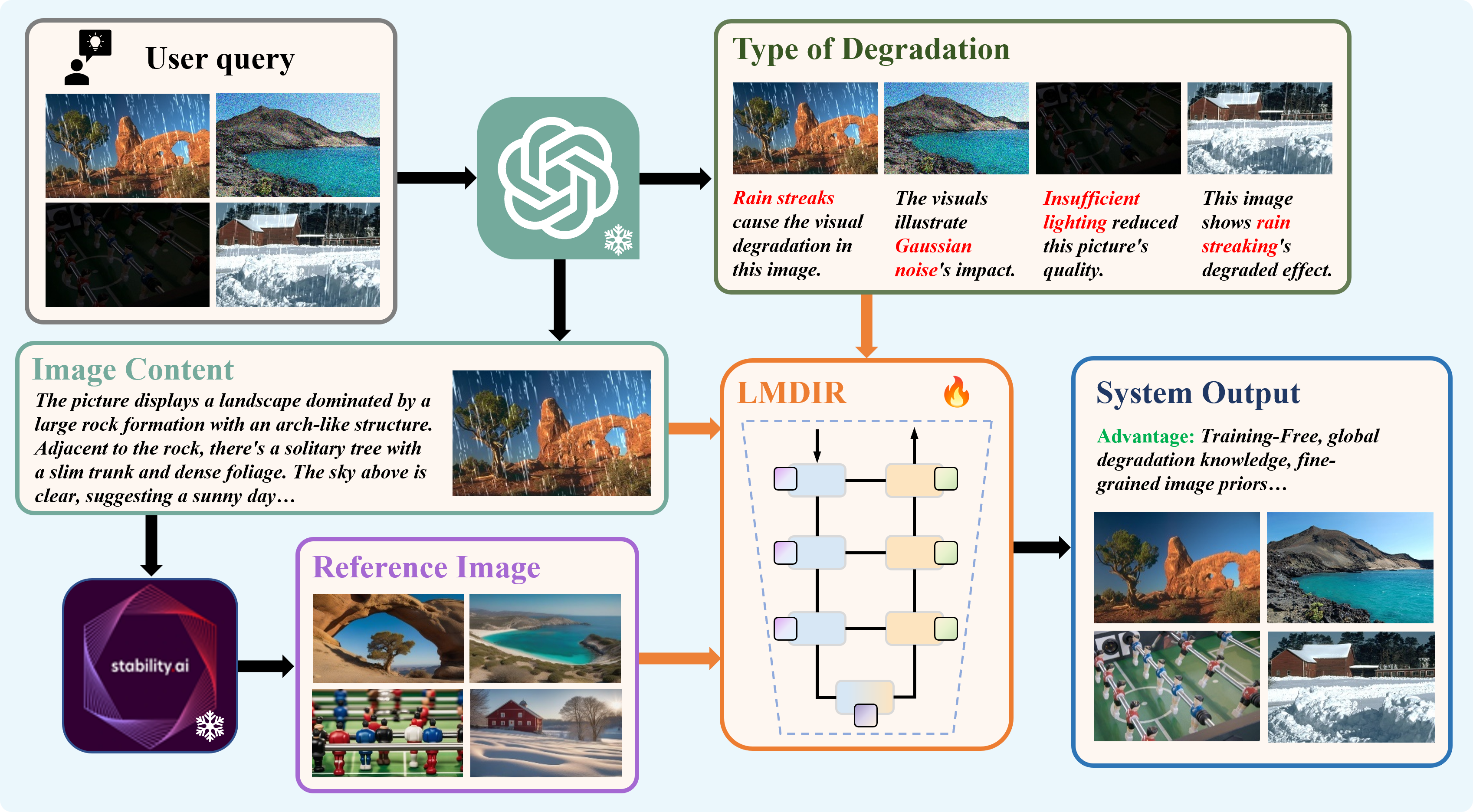}
    \caption{The overall pipeline of our proposed method. It achieves high-quality multiple-in-one image restoration with large model prior.}
    \label{fig:motivation}
\end{figure*}
Recently, the image restoration community has shifted its focus toward multiple-in-one restoration tasks~\cite{li2022all,potlapalli2023promptir,luo2023controlling}, where a single model is tasked with handling multiple degradation types. This multi-task capability is achieved by injecting degradation-relevant knowledge into the model, enabling it to discriminate between different degradation types and process image features dynamically. The performance of such models heavily relies on the accurate perception of degradation embeddings~\cite{kong2024towards}. 
A pioneering work in this task, AirNet~\cite{li2022all}, learns explicit degradation representations through contrastive learning, while DA-CLIP~\cite{luo2023controlling} generates accurate embeddings by fine-tuning a pre-trained CLIP~\cite{clip} model. In contrast to explicit embedding methods, PromptIR~\cite{potlapalli2023promptir} and ProRes~\cite{ma2023prores} utilize prompt learning for implicit embedding learning. However, the former explicit embedding methods typically require a two-stage training approach, consuming large computational resources, especially when fine-tuning large pre-trained models. On the other hand, the latter implicit embedding approaches struggle to generate accurate representations, and the training process itself can be challenging~\cite{kong2024towards}.

The recent emergence of multi-modal large language models (MMLM)~\cite{yin2023survey} offers a promising solution to address these challenges. These models, trained on large-scale image-text-paired datasets, possess strong capabilities in image captioning, visual question answering, and scene understanding. Notably, they have demonstrated a powerful ability to comprehend low-level image features, as evidenced by their performance on the Q-Bench~\cite{wu2023q} benchmark. Leveraging this understanding, MMLMs can vividly and accurately describe image degradations and contents, providing reliable prior information to restoration models without the need for complex fine-tuning or multi-stage training procedures.
In addition to the global prior derived from textual descriptions, local fine-grained priors obtained from reference images can further enhance the performance of restoration models. This approach has been explored in the context of reference-based super-resolution tasks~\cite{jiang2021robust}. Leveraging the powerful generative capabilities of state-of-the-art diffusion models~\cite{podell2023sdxl}, we can synthesize high-quality reference images that share similar content and semantic context with the input degraded image. These reference images are generated in a guided manner, informed by the contextual text descriptions produced by the multi-modal language models.

Motivated by the observations discussed earlier, we propose a novel multiple-in-one image restoration framework, dubbed LMDIR (Large Model Driven Image Restoration Framework), that leverages prior knowledge from large multi-modal models to tackle diverse image degradations. As illustrated in Figure~\ref{fig:motivation}, LMDIR incorporates three essential priors: 1) global degradation knowledge derived from MMLMs; 2) scene-aware contextual descriptions generated by MMLMs; and 3) fine-grained high-quality reference images synthesized by diffusion models guided by the MMLM-generated contextual descriptions. Building upon these priors, the proposed LMDIR architecture consists of four main components: a customized query-based prompt encoder that refines textual information from MMLMs by leveraging image low-level features, a degradation-aware transformer block that incorporates global degradation knowledge, a content-aware transformer block that utilizes the scene-aware content descriptions, and a reference-based transformer block that integrates fine-grained image priors from the synthesized reference images through global and local perspectives. This design empowers LMDIR to adopt a single-stage training strategy capable of addressing diverse and complex image restoration tasks, while also offering the flexibility of automatic or user-guided restoration based on provided prompts. Extensive experiments validate the superiority of LMDIR over other state-of-the-art multiple-in-one image restoration methods across multiple evaluation benchmarks.

Our key contributions can be summarized as follows:
\begin{enumerate}
    \item We introduce LMDIR, an innovative framework that harnesses the capabilities of multi-modal large language models and diffusion models to address the challenges of multiple-in-one image restoration.Extensive experiments have shown that LMDIR outperforms state-of-the-art methods for multiple-in-one image restoration tasks.
    \item We introduced a query-based prompt encoder that refines text from multi-modal large language models (MMLMs), enabling automatic or user-guided restoration. We also designed degradation-aware transformer blocks to incorporate global degradation knowledge, enhancing the model's capability to handle diverse types of degradations. Additionally, we utilized reference-based transformer blocks that leverage fine-grained image priors from synthesized reference images, further improving the quality of image restoration. 
    \item Through extensive experiments, we demonstrate that LMDIR outperforms existing state-of-the-art methods on multiple evaluation metrics for multiple-in-one image restoration tasks.
\end{enumerate}

\section{Related Work}
\subsection{Multiple-in-one Image Restoration}
Image restoration endeavors to reconstruct high-quality images from degraded versions that have been impacted by a range of degradations, including noise \cite{li2023ntire,chen2023masked}, rain \cite{rain1,rain2}, low-light conditions \cite{zerodce,ren2019low}, and other factors \cite{desnowing, deblurring}. Each degradation type exhibits unique characteristics and introduces distinct distortions during the imaging process. Consequently, previous studies have predominantly focused on designing specialized models tailored to handle specific restoration tasks by leveraging prior knowledge about the respective degradations. However, this approach limits the applicability of such models in real-world scenarios, where the degradation type can be dynamic and time-varying.
Recently, the image restoration community has shifted its attention towards multiple-in-one restoration tasks, which involve developing a single model capable of handling various types of degradation. Early attempts in this direction employed networks with multiple encoders and decoders, where different encoder-decoder pairs were dedicated to specific degradation types~\cite{chen2021pre,li2020all}. However, these methods required prior knowledge of the degradation type and were primarily focused on addressing adverse weather conditions.
AirNet~\cite{li2022all} introduced a two-stage training approach that combined an explicit degradation classifier with contrastive learning to adaptively recognize degradation types and simultaneously perform denoising, rain removal, and image dehazing tasks. Subsequently, PromptIR~\cite{potlapalli2023promptir} and ProRes~\cite{ma2023prores} leveraged prompt learning techniques~\cite{zhou2022conditional} to achieve implicit degradation representation learning, eliminating the need for separate degradation classifiers and two-stage training. DA-CLIP~\cite{luo2023controlling}, on the other hand, fine-tuned a pre-trained CLIP model to generate accurate degradation embeddings, which were then injected into the restoration network.
Explicit classification methods in image restoration typically rely on computationally expensive two-stage training procedures. Further, implicit prompt learning methods often encounter difficulties in generating accurate representations of degradations, and the training process itself can be challenging.

\subsection{Text Driven Image Manipulation}
In recent years, significant advancements have been made in text-based image generation and editing~\cite{lee2021brief}. VQGAN-CLIP~\cite{crowson2022vqgan} combines pre-trained generative models and CLIP to guide the generation process toward a desired target description. Additionally, latent diffusion models~\cite{rombach2022high} have been introduced, which can effectively follow user instructions and improve image quality through text guidance.

Beyond image generation, some tasks have also explored user-guided image editing and painting, such as InstructPix2Pix~\cite{brooks2023instructpix2pix} and Imagic~\cite{kawar2023imagic}.
The progress of diffusion models led to the development of more sophisticated models such as Emu Edit~\cite{sheynin2024emu}. This approach not only processes standard image inputs but also incorporates depth maps. Concurrently, methods like LEDITS++~\cite{brack2024ledits++} have pushed the boundaries of image generation fidelity by leveraging the power of DDPM inversion. 
However, the field of image restoration has not fully explored the potential of text-driven image restoration.
\begin{figure*}[!h]
    \centering
    \includegraphics[width=\linewidth]{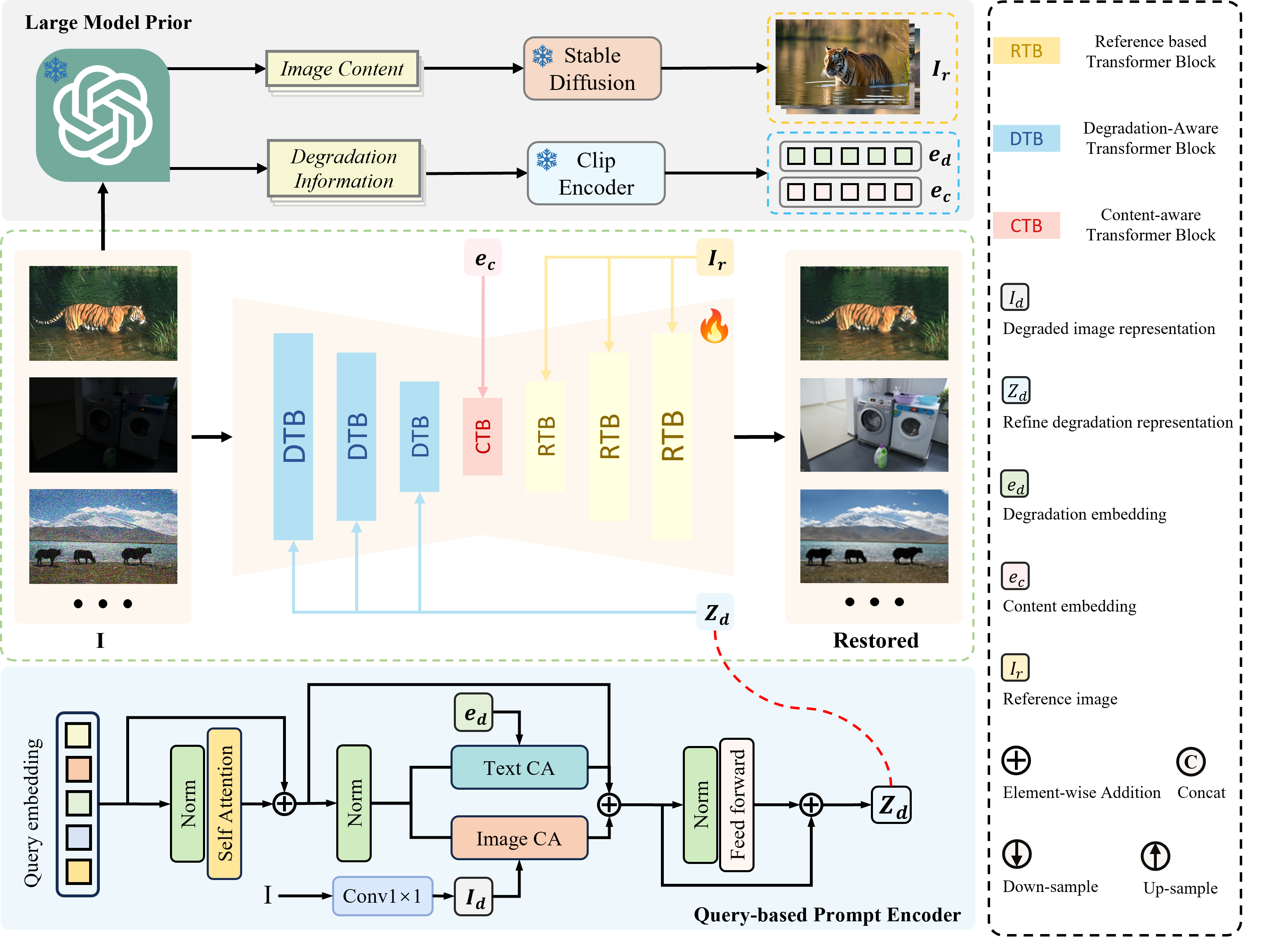}
    \caption{The framework of our network. We utilized the pretrained MLLM, CLIP and diffusion models for generating the prior information to guide the restoration process.}
    \label{fig:arch}
\end{figure*}
\subsection{Reference-based Image Super-Resolution}
Another area related to our work is reference-based super-resolution, which differs from single image super-resolution in that it leverages reference images to assist in the super-resolution process by extracting similar textures and details. The reference images are highly similar to the groundtruth high-resolution images.
Representative works in this field including CrossNet~\cite{zheng2018crossnet}, for instance, establishes inter-image correlations by estimating optical flow between reference and low-resolution images. SRNTT~\cite{zhang2019image} computes similarities between these images and transfers textures from the reference to enhance the low-resolution counterparts. Furthermore, TTSR~\cite{yang2020learning} introduces both hard and soft attention mechanisms to facilitate texture transfer and synthesis. Lastly, C2-Matching~\cite{jiang2021robust} pioneers the use of a contrasting correlation network to learn image correlations, followed by a teacher-student correlation distribution to refine the alignment between low-resolution and high-resolution images, thereby enhancing the overall quality of super-resolved images.
While significant progress has been made in the reference-based super-resolution field, these methods require manual selection of reference images by users. In contrast, our approach leverages a diffusion model to adaptively generate reference images with highly similar content to the ground truth image, and these generated reference images are then used to improve the performance of image restoration models.
\section{Method}
In this section, we first introduce the three large model priors utilized, followed by a detailed description of the proposed framework.
\subsection{Large Model Prior}
\subsubsection{Global degradation and content prior}
Unlike previous methods that require fine-tuning on large models or two-stage training, we obtain content and degradation embeddings in a training-free manner. We leverage prompt engineering to query a multimodal language model, which outputs degradation information present in the image as well as content information unrelated to degradation. We then obtain the corresponding embeddings of the generated text using the CLIP encoder, serving as global degradation and content embeddings to guide the model's training. We utilize the GPT4o model~\cite{gpt4o}, which has demonstrated strong performance in low-level tasks, to generate this global degradation priors $\mathbf{e}_d$ and content text embedding $\mathbf{e}_c$, shown in the top of Figure \ref{fig:arch}.

\subsubsection{Local content prior}
In addition to the global prior knowledge provided by text, we also utilize images generated by the diffusion model as fine-grained content priors, providing detailed texture and feature references for image restoration models. Specifically, we input the content text output by the multimodal large language model into the SDXL~\cite{podell2023sdxl} model as a prompt and use the degraded text as a negative prompt, ensuring that the generated image shares similar content with the ground truth.

\subsection{Model Architecture}
Figure~\ref{fig:arch} illustrates the overall framework of our proposed method, comprising an image restoration network, a query-based prompt encoder, a multi-modal language model, a diffusion model, and a CLIP encoder. Given a degraded input image $\mathbf{I}\in \mathbb{R}^{\rm H\times W\times 3}$, we first pass it and a prompt text through the multi-modal language model (MLLM) to generate a degradation text embedding $\mathbf{T}_d$ and a content text embedding $\mathbf{T}_c$, respectively. These text embeddings are then encoded by the CLIP encoder to obtain a degradation embedding $\mathbf{e}_d \in \mathbb{R}^{\rm N \times C}$ and a content embedding $\mathbf{e}_c \in \mathbb{R}^{\rm N\times C}$. Concurrently, the input image $\mathbf{I}$ is processed by a simple image encoder built on residual blocks to obtain a degraded image representation $\mathbf{I}_d \in \mathbb{R}^{\rm C}$.

We feed the degradation encoding $\mathbf{e}_d$ and the identity encoding $\mathbf{I}_d$ into the query-based prompt encoder to refine the degradation representation as $\mathbf{Z}_d \in \mathbb{R}^{\rm N\times C}$. Concurrently, we input the content encoding $\mathbf{T}_c$ to the diffusion model to synthesize a high-quality reference image $\mathbf{I}_r \in \mathbb{R}^{\rm H\times W\times 3}$. Finally, the backbone restoration network takes the refined degradation representation $\mathbf{Z}_d$, the content encoding $\mathbf{e}_c$, and the reference image $\mathbf{I}_r$ as conditions to restore the output image $\mathbf{Y} \in \mathbb{R}^{\rm H\times W\times 3}$ from the degraded input $\mathbf{I}$.

Our framework effectively integrates global degradation priors $\mathbf{T}_d$ and scene-aware content priors $\mathbf{T}_c$ extracted from the multi-modal language model (MLLM), as well as fine-grained reference priors $\mathbf{I}_r$ generated by the diffusion model. This integrated approach enables robust multiple-in-one image restoration capabilities, leveraging complementary information from the language and diffusion models to tackle a variety of image degradation challenges.

\subsection{Key Components}
\subsubsection{Query-based Prompt Encoder}
The degradation embedding $\mathbf{e}_d$ extracted directly from the CLIP encoder cannot be directly applied as a prior for the image restoration network due to two reasons: 1) The CLIP encoder lacks awareness of specific degradation details such as rain streaks and noise distribution, providing only global classification knowledge. 2) The textual description generated by the multi-modal language model may not be entirely reliable. Therefore, we design a query-based prompt encoder to refine $\mathbf{e}_d$ into a more fine-grained degradation representation $\mathbf{Z}_d$ that can effectively guide the restoration network, while incorporating degradation information from the image itself.
Specifically, given a learnable query embedding $\mathbf{E}_p \in \mathbb{R}^{\rm \hat{N}\times C}$, the degradation text embedding $\mathbf{e}_d$ from CLIP, and the degraded image representation $\mathbf{I}_d$, the query-based prompt encoder computes the refined degradation representation $\mathbf{Z}_d$. In detail, $\mathbf{E}_p$ attends to itself via $\textrm{SA}(.)$ to obtain $\mathbf{E}_p'$, which is projected to queries $\mathbf{Q}_{E_p}$. Then, cross-attention is performed between $\mathbf{Q}_{E_p}$ and keys/values from $\mathbf{e}_d$ to obtain $\mathbf{Z}_\text{text}$ encoding degradation information from text, and with $\mathbf{I}_d$ to obtain $\mathbf{Z}_\text{image}$ encoding image degradation information as:
\begin{align}
&\mathbf{E}_p' = \textrm{SA}(\mathbf{E}_p), \\
&\mathbf{Q}_{E_p} = \mathbf{E}_p'W_{qp}, \\
&\mathbf{K}_{e_d}, \mathbf{V}_{e_d} = \mathbf{e}_dW_{kd}, \mathbf{e}_dW_{vd}, \\
&\mathbf{K}_{I_d}, \mathbf{V}_{I_d} = \mathbf{I}_dW_{ki}, \mathbf{I}_dW_{vi},
\end{align}
where $\textrm{SA}(.)$ and $\textrm{CA}(.)$ denote self-attention and cross-attention, respectively. Finally, $\mathbf{Z}_\text{text}$ and $\mathbf{Z}_\text{image}$ are fused and processed by a feed-forward network (FFN)~\cite{geva2020transformer} to yield the refined degradation representation $Z_d$ as
\begin{align}
&\mathbf{Z}_\text{text} = \textrm{CA}(\mathbf{Q}_{E_p}, \mathbf{K}_{e_d}, \mathbf{V}_{e_d}), \\
&\mathbf{Z}_\text{image} = \textrm{CA}(\mathbf{Q}_{E_p}, \mathbf{K}_{I_d}, \mathbf{V}_{I_d}), \\
&\mathbf{Z}_d = \textrm{FFN}(\mathbf{Z}_\text{text} + \mathbf{Z}_\text{image}).
\end{align}
This representation $\mathbf{Z}_d$ combines the information from text prior and the image feature, can provide restoration network with a better degradation presentation. 
\begin{figure}
    \centering
\includegraphics[width=\linewidth]{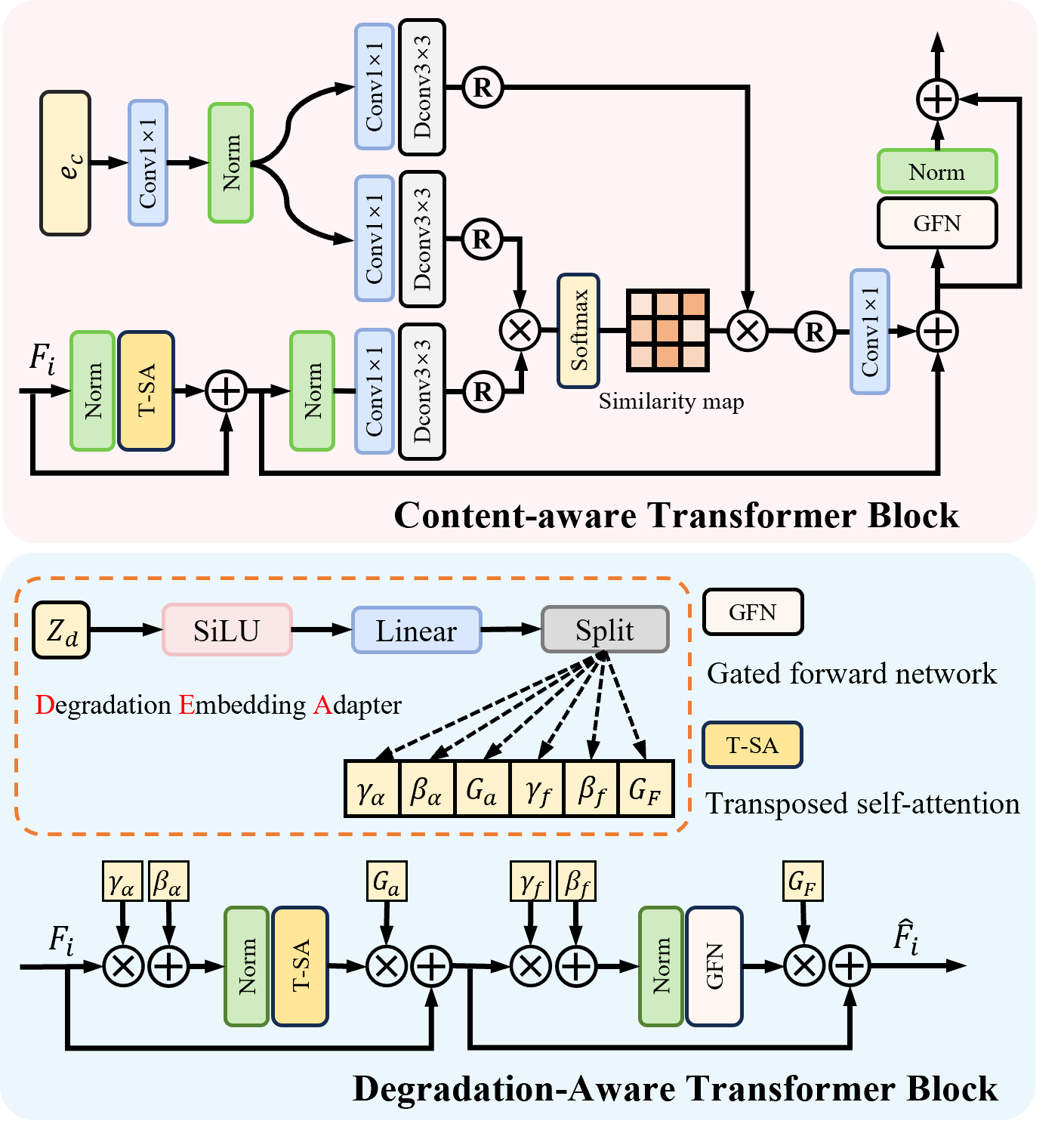}
    \caption{Our proposed content aware Tranformer block and degradation aware transformer block. These blocks are utilized to inject prior knowledge from content and degradation prior.}
    \label{fig:degblock}
\end{figure}
\subsubsection{Degradation-Aware Transformer Block}
In the encoder part of our image restoration model, we employ degradation-aware transformer blocks to inject degradation information and enable dynamic feature processing based on the degradation type. Specifically, each degradation-aware transformer block consists of three components: transposed self-attention~\cite{zamir2022restormer}, gated feed-forward network~\cite{zamir2022restormer}, and a degradation embedding adapter, as shown in Figure.~\ref{fig:degblock}.
Given the input feature map $\mathbf{F_i}$ and degradation embedding $\mathbf{Z}_d$, the operations are defined as follows:
\begin{align}
&\mathcal{G_A},\mathcal{G_F},\gamma_a, \beta_a,\gamma_f, \beta_f = \textrm{DEA}(\mathbf{Z}_d),\\
&\tilde{F}_i = \mathcal{G_A}\odot\textrm{TSA}(\gamma_a\odot F_i+\beta_a)+F_i, \\
&\hat{F}_i = \mathcal{G_F}\odot\textrm{GFN}(\gamma_f \odot \tilde{F}_i+\beta_f)+\tilde{F}_i
\end{align}
where transposed self-attention $\textrm{TSA}(.)$ captures long-range dependencies in $F_i$, Gated forward network $\textrm{GFN}(.)$ refines the local feature. The degradation embedding adapter $\textrm{DEA}(.)$ projects $\mathbf{Z}_d$ to the same channel dimension as $\hat{F}_i$ and further generating degradation-aware parameters.
Specifically, degradation adapter generate the degradation-aware parameters using:
\begin{align}
    &\tilde{Z}_d = \text{SiLU}(W_{\text{adapt}} Z_d), \\
    &E = W_{\text{linear}} \tilde{Z}_d,\\
    &(\mathcal{G_A}, \mathcal{G_F}, \gamma_a, \beta_a, \gamma_f, \beta_f) = \text{split}(E, 6).
\end{align}
where $\gamma$ and $\beta$ and $
\mathcal{G}$ are scale, shift and gate parameters modulated by $\mathbf{Z}_d$. $Split(.)$ is the split operator along channel dimension.
By integrating $\mathbf{Z}_d$ into the transformer blocks, the model can dynamically adapt its feature processing based on specific degradation representation, enabling effective restoration for diverse degradations using a single model.

\subsubsection{Content-aware Transformer Block}
In the bottleneck parts of our image restoration network, we designed content-aware transformer blocks to incorporate local content features and enhance restoration performance. In the bottleneck, we utilize the content text embedding $\mathbf{e}_c$ as a reference. As shown in Figure~\ref{fig:degblock}.
We first project $\mathbf{e}_c$ to the same dimension as the feature map $\mathbf{F}_i$ using a multi-layer perceptron. Then, we perform self-attention on $\mathbf{F}_i$ and calculate the similarity between $\mathbf{F}_i$ and the projected $\mathbf{e}_c$. Based on this similarity, we adaptively select and integrate useful features from $\mathbf{e}_c$ into $\mathbf{F}_i$, followed by a gated FFN for local feature processing. This operation injects global content priors from the text embedding into the network.
The content-aware transformer block can be formulated as:
\begin{align}
&\tilde{\mathbf{F}}_i = \textrm{TSA}(\mathbf{F}_i)+\mathbf{F}_i, \\
&\hat{\mathbf{F}}_i = \textrm{RA}(\tilde{\mathbf{F}}_i, \mathbf{e}_c)+\tilde{\mathbf{F}}_i, \\
&\mathbf{F}_{i+1} = \textrm{GFN}(\hat{\mathbf{F}}_i).
\end{align}
Here we utilized the reference-attention $RA(.)$ to inject the reference feature, due to the token length of $\mathbf{e}_c$ is a fixed number. The integrated features $\hat{\mathbf{F}}_i$ are further processed by a gated FFN to produce the output $\mathbf{F}_{i+1}$.
Given the input feature $\tilde{\mathbf{F}}_i$ and the reference feature denoted as $\mathbf{e}_c$, the operation of $\textrm{RA}(.)$ can be defined as follows:
\begin{align}
    &Q = \tilde{\mathbf{F}}_i W_q,\\
    &K,V = \mathbf{F}_{Ref}W_k, \mathbf{F}_{Ref}W_v,\\
    &Sim  = \texttt{softmax}(QK^{T}),\\
    &\mathbf{F}_{out} = Sim*V.
\end{align}
\begin{figure}
    \centering
\includegraphics[width=\linewidth]{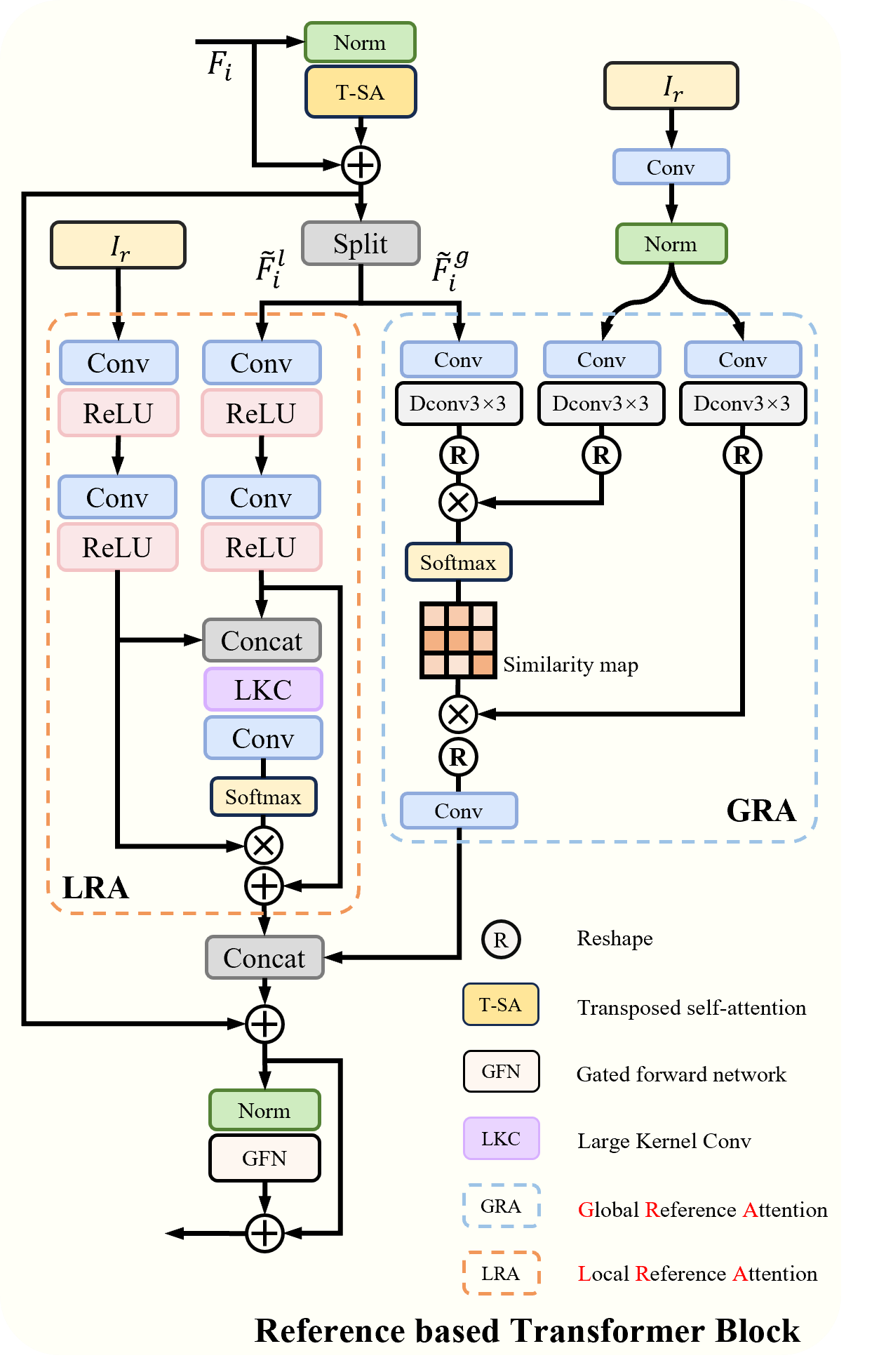}
    \caption{Our proposed reference based Tranformer block. This block incorporates details from reference image through local and global reference attention.}
    \label{fig:refblock}
\end{figure}
\subsubsection{Reference-based Transformer Block}
In the decoder parts of our image restoration network, we introduce reference-based transformer blocks to integrate fine-grained reference features~\ref{fig:refblock}. Specifically, we leverage the reference image $\mathbf{I}_r$, generated by the diffusion model, as our reference. This block is designed to extract both global and local similar features from the reference image. To achieve this, we employ a hybrid approach that combines global and local attention mechanisms. The global reference attention utilizes a transposed cross-attention mechanism to compute the similarity between the two images along the channel dimension. In contrast, the local reference attention employs convolution to fuse similarity features along the spatial dimension.
Given the input feature $\mathbf{F}_i$ and reference image $\mathbf{I}_r$, this process can be described as follows:
\begin{align}
    &\mathbf{F}_{ref} = \phi(\mathbf{I}_r),\\
    &\tilde{\mathbf{F}}_i = \textrm{TSA}(\mathbf{F}_i) + \mathbf{F}_i, \\
    &\tilde{\mathbf{F}}^l_i, \tilde{\mathbf{F}}^g_i = \textrm{split}(\tilde{\mathbf{F}}_i, 2),\\
    &\hat{\mathbf{F}}_i = \Theta([\textrm{LRA}(\tilde{\mathbf{F}}^l_i, \mathbf{F}_{ref}), \textrm{GRA}(\tilde{\mathbf{F}}^g_i, \mathbf{F}_{ref})]) + \tilde{\mathbf{F}}_i, \\
    &\mathbf{F}_{i+1} = \textrm{GFN}(\hat{\mathbf{F}}_i)+\hat{\mathbf{F}}_i.
\end{align}
Here, $\phi(.)$ is the convolution operator that projects $\mathbf{I}_r$ to $\mathbf{F}_{ref}$ for dimension alignment. The $\textrm{TSA}(.)$ and $\textrm{split}(.)$ are the transposed self-attention and channel split operators, respectively. After generating the outputs from local reference attention $\textrm{LRA}(.)$ and global reference attention $\textrm{GRA}(.)$, the two features are concatenated and fused through the linear projection $\Theta(.)$. Finally, a gated forward network, $\textrm{GFN}(.)$, is utilized to enhance the locality of the features.
The $\textrm{GRA}(.)$ is the cross attention version of $\textrm{TSA}(.)$, where $Q$ is derived from $\tilde{\mathbf{F}}_i$ and $K,V$ are generated from $\mathbf{F}_{ref}$. The local reference attention can be described as below:
\begin{align} 
&\mathbf{F}_j =\mathbf{W}_2 * \text{ReLU}(\mathbf{W}_1 * \tilde{\mathbf{F}}^l_i) \\ &\mathbf{F}_k = \mathbf{W}_2 * \text{ReLU}(\mathbf{W}_1 * \mathbf{F}{ref})\\ 
& Sim = \text{Softmax}(\mathbf{W}_a * (\mathbf{F}_j + \mathbf{F}_k)) \\
&\mathbf{F}_{\text{agg}} = \mathbf{F}_j + Sim \odot \mathbf{F}_k 
\end{align}
where ($\mathbf{W}_1$), ($\mathbf{W}_2$), and ($\mathbf{W}_a$) are convolutional filters, ( $*$ ) denotes convolution operation, ($\odot$) denotes element-wise multiplication, and ($\text{ReLU}$) and ($\text{Softmax}$) are the activation function and softmax function, respectively.

\subsection{Loss Function}
Following the widely-adapted methods, we utilized L1 norm between the output $Y$ and groundtruth $G$ as our loss function:
\begin{equation}
    L = ||Y-G||_1
\end{equation}

\begin{table*}[!ht]
\centering
\Large
 	\renewcommand{\tabcolsep}{2pt} 
\renewcommand{\arraystretch}{1.5}
\caption{Quantitative comparison of our method with other state-of-the-art approaches in noise-rain-lowlight settings. PSNR/SSIM values are reported. The best results are marked in bold.}
\label{table:maintable}
\resizebox{\linewidth}{!}{
\begin{tabular}{c|ccc|ccc|c|c|c}
\hline
\multirow{2}{*}{Method} & \multicolumn{3}{c|}{Denoise(BSD68)}        & \multicolumn{3}{c|}{Denoise(Urban100)}     & \multirow{2}{*}{Derain} & \multirow{2}{*}{lowlight} & \multirow{2}{*}{Average} \\
                        & $\sigma=15$           & $\sigma=25$          & $\sigma=50$            & $\sigma=15$           & $\sigma=25$           & $\sigma=50$           &                         &                                              &                          \\ \hline
HINet                   & 32.35/0.925  & 26.09/0.869 & 25.91/0.767   & 33.68/0.938  & 30.63/0.908  & 27.50/0.850  & 37.63/0.980             & 16.55/0.769                                  & 28.79/0.875              \\
NAFNet                  & 32.93/0.915  & 30.36/0.862 & 27.22/0.759  & 31.98/0.920 & 29.56/0.881 & 26.24/0.795  & 32.22/0.939             & 20.72/0.777                                 & 28.90/0.856              \\
SwinIR                  & 33.62/0.926  & 31.00/0.879 & 27.68/0.780   & 33.57/0.938  & 31.13/0.906  & 27.60/0.835  & 34.32/0.965             & 18.86/0.800                                & 29.72/0.878              \\
Restormer                  & 33.67/0.924  & 31.07/0.876 & 27.86/0.782   & 33.46/0.934  & 31.09/0.904  & 27.80/0.837  & 36.55/0.974             & 21.49/0.822                                  & 30.37/0.881              \\
AirNet                  & 33.66/0.923  & 31.10/0.881 & 27.72/0.780   & 33.55/0.937 & 31.10/0.905 & 27.77/0.837 & 35.80/0.971            & 16.21/0.673                                  & 29.61/0.863              \\
PromptIR                & 33.63/0.927  & 31.02/0.880 & 27.77/0.782   & 33.45/0.937  & 31.05/0.907  & 27.71/0.839  & 36.37/0.975             & 21.14/0.831                                  & 30.27/0.884              \\
DA-CLIP                 & 30.30/0.837 & 27.54/0.758 & 24.77/0.619 & 29.30/0.819   & 25.18/0.634  & 23.71/0.613 & 36.37/0.965             & 19.06/0.789                                  & 27.03/0.754              \\ \hline
Ours                    & \textbf{34.00}/\textbf{0.930}  & \textbf{31.38}/\textbf{0.886} & \textbf{28.15}/\textbf{0.798}   & \textbf{34.15}/\textbf{0.945}  & \textbf{31.84}/\textbf{0.919}  & \textbf{28.62}/\textbf{0.873} & \textbf{38.64}/\textbf{0.983}             & \textbf{23.24}/\textbf{0.850}                                  & \textbf{31.25}/\textbf{0.898}              \\ \hline
\end{tabular}}
\end{table*}
\begin{figure*}[!h]
    \centering
    \includegraphics[width=\linewidth]{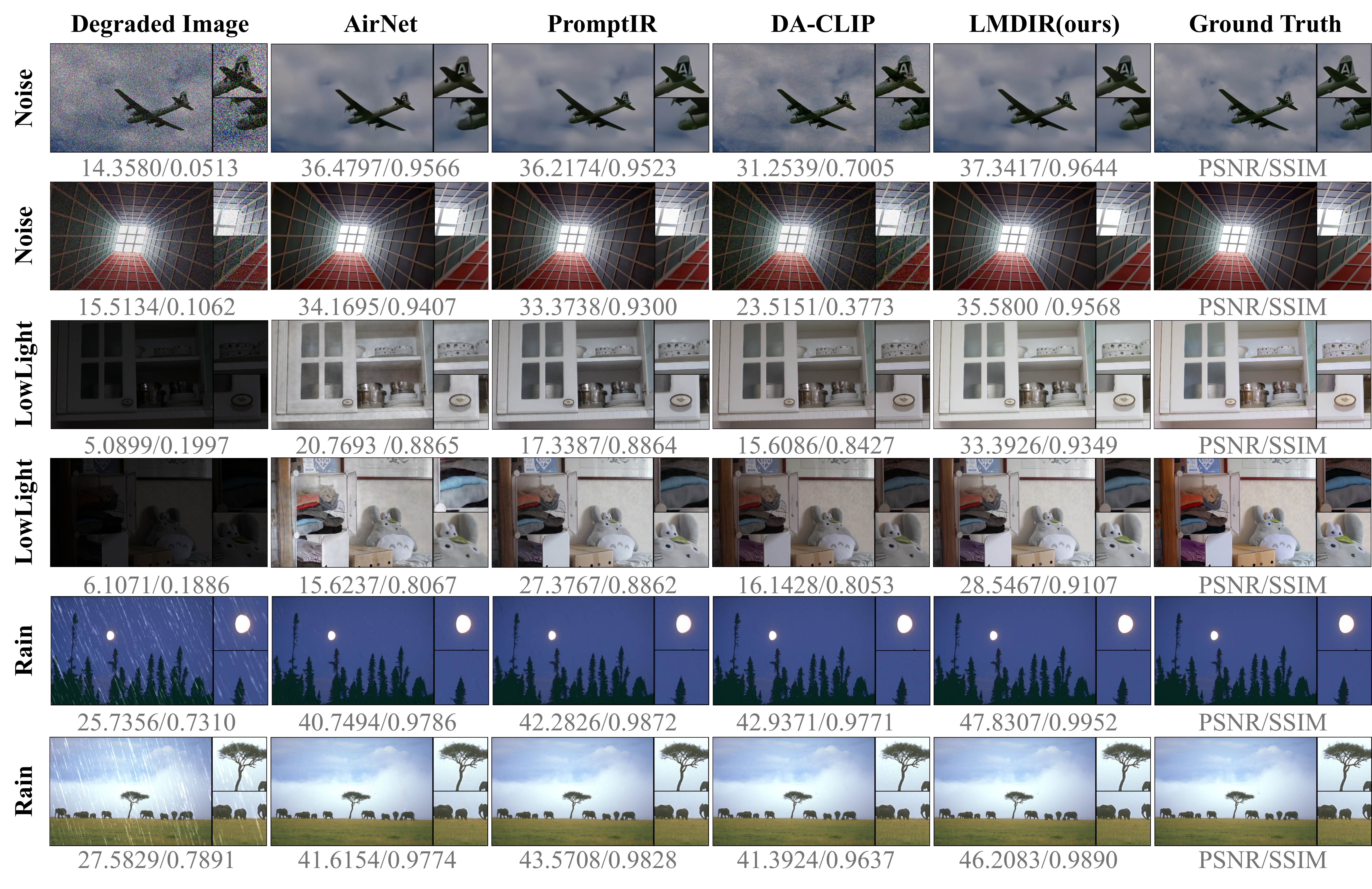}
    \caption{Visual comparison of multiple-in-one methods on image denoising, low light enhancement, and deraining.}
    \label{fig:visualresults}
\end{figure*}
\begin{figure*}[!h]
    \centering
    \includegraphics[width=\linewidth]{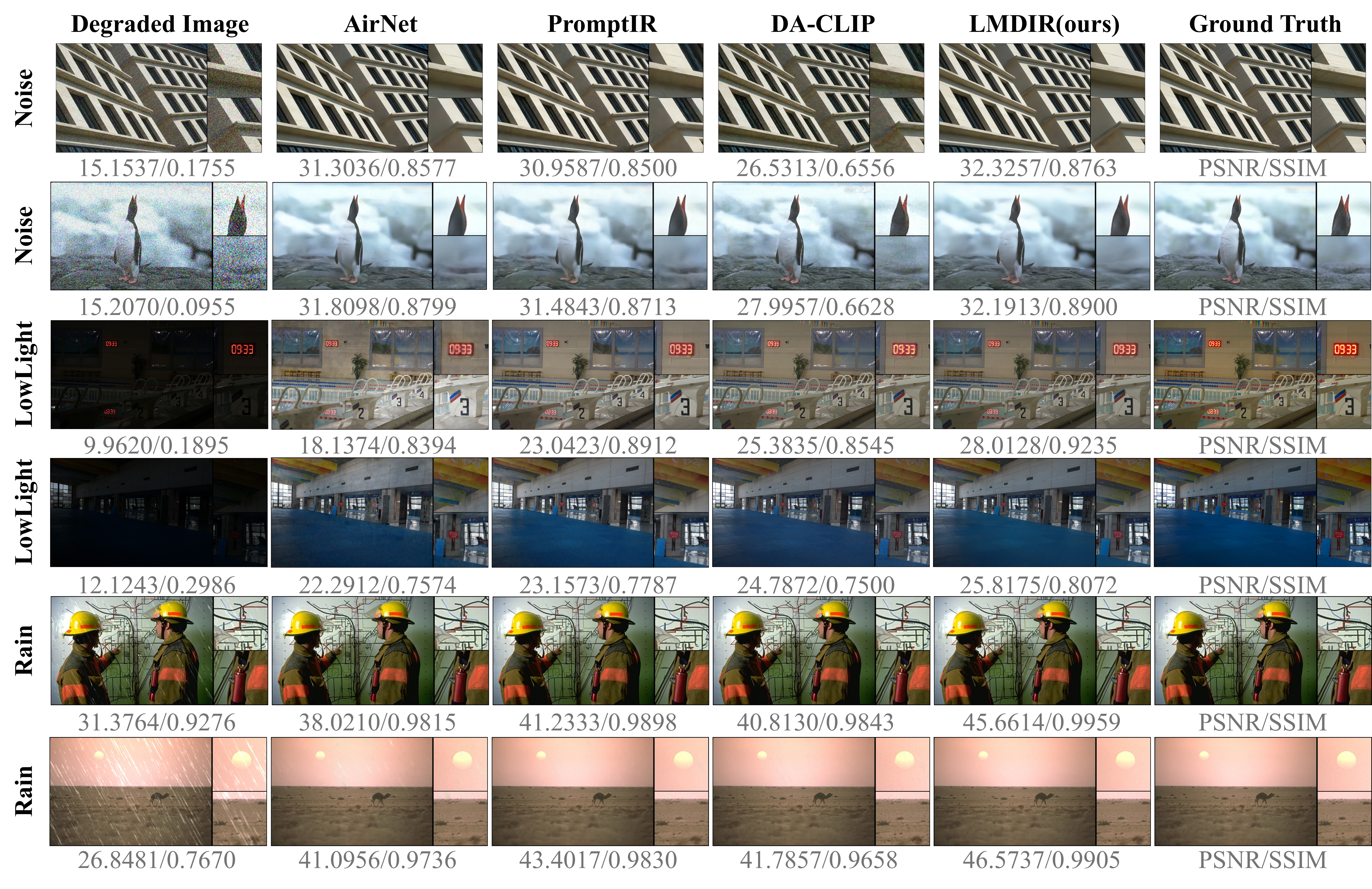}
    \caption{Visual comparison of multiple-in-one methods on image denoising, low light enhancement, and deraining.}
    \label{fig:visualresults2}
\end{figure*}
\section{Experiments}
\subsection{Datasets and Benchmark}
We evaluate our method on a multiple-in-one image restoration task comprising three representative subtasks: image deraining, image denoising, and low-light image enhancement.
For image deraining datasets, we chose the Rain1800~\cite{rain1800} dataset for training and evaluate on 100 test images from the Rain100L~\cite{yang2017deep} dataset. For denoising, we use synthetically generated noisy image with noise level of $\sigma \in \{15,25,50\}$ on the WED~\cite{wed} dataset for training, and evaluate on the Urban100~\cite{huang2015single} and BSD68~\cite{bsd68} datasets. For low-light enhancement, we train on the LOL~\cite{wei1808deep} dataset and test on its corresponding test set. During training, we randomly sample these three datasets with a uniform distribution.
We compare our method against classic image restoration networks (HINet~\cite{chen2021hinet}, NAFNet~\cite{chen2022simple}, SwinIR~\cite{liang2021swinir},
Restormer~\cite{zamir2022restormer})
and recent multiple-in-one approaches (AirNet~\cite{li2022all}, PromptIR~\cite{potlapalli2023promptir}, DA-CLIP~\cite{luo2023controlling}).
We adopt PSNR and SSIM to assess the performance of model.
\subsection{Implementation Details}
We train our model using the PyTorch framework on a single NVIDIA RTX 3090 GPU with the Adam optimizer. During training, images are randomly cropped into 128×128 patches with a batch size of 2. The total number of training iterations is 300000. The initial learning rate is set to 2e-4 for the whole training process.

To generate degradation text $\mathbf{T_d}$ and content text $\mathbf{T_c}$, we utilize the GPT4o multi-modal language model. For synthesizing reference images $\mathbf{I_r}$, we employ the Stable Diffusion XL (SDXL) v1.0 diffusion model with 30 sampling steps. We generate all the reference image and text descriptions before training our model.

To ensure a fair comparison, we retrain all baseline models using the same framework from PromptIR~\cite{potlapalli2023promptir} and identical hyperparameters.

\subsection{Comparison with Sota Methods}
\subsubsection{Multiple-in-one restoration evaluation} In Table~\ref{table:maintable}, we present a comparison between our proposed LMDIR approach and existing state-of-the-art methods, demonstrating substantial enhancements across various tasks. Notably, in comparison to PromptIR, our method achieved an average improvement of 2.3 dB in PSNR of the image deraining task. Additionally, the denoising and low-light image enhancement tasks exhibited marked progress. PromptIR's inability to produce accurate restoration outcomes can be attributed to its implicit degenerate feature learning method. It is noteworthy to highlight that DA-CLIP necessitates an extensive volume of data for training due to its reliance on diffusion models and fails to yield satisfactory results within our settings. Contrasting with these methods, our approach leverages the prior knowledge provided by the large model and the information intrinsically present in the degraded image, resulting in superior performance.

The comparative visualization of different methods is shown in Figures~\ref{fig:visualresults} and~\ref{fig:visualresults2}. For each task, we opted for two representive images for visual comparison. Within the the denoising task, we set the noise level $\sigma$=50 for comparison. As the figure depicted, our methods outperforms others in achieving superior restoration outcomes. In the context of denoising, DA-CLIP falls short in complete noise reducing, whereas PromptIR induces a loss of high-frequency details within the image. In the low-light image enhancement task, the color accuracy of our method closely aligns with the ground truth, while the results generated by AirNet manifest a dark texture. Evaluating the image deraining results, residual rain streaks are discernible in the images produced by AirNet and PromptIR. In contrast, our method exhibits the highest quality in rain removal results.
\begin{figure*}[h]
    \centering
    \includegraphics[width=\linewidth]{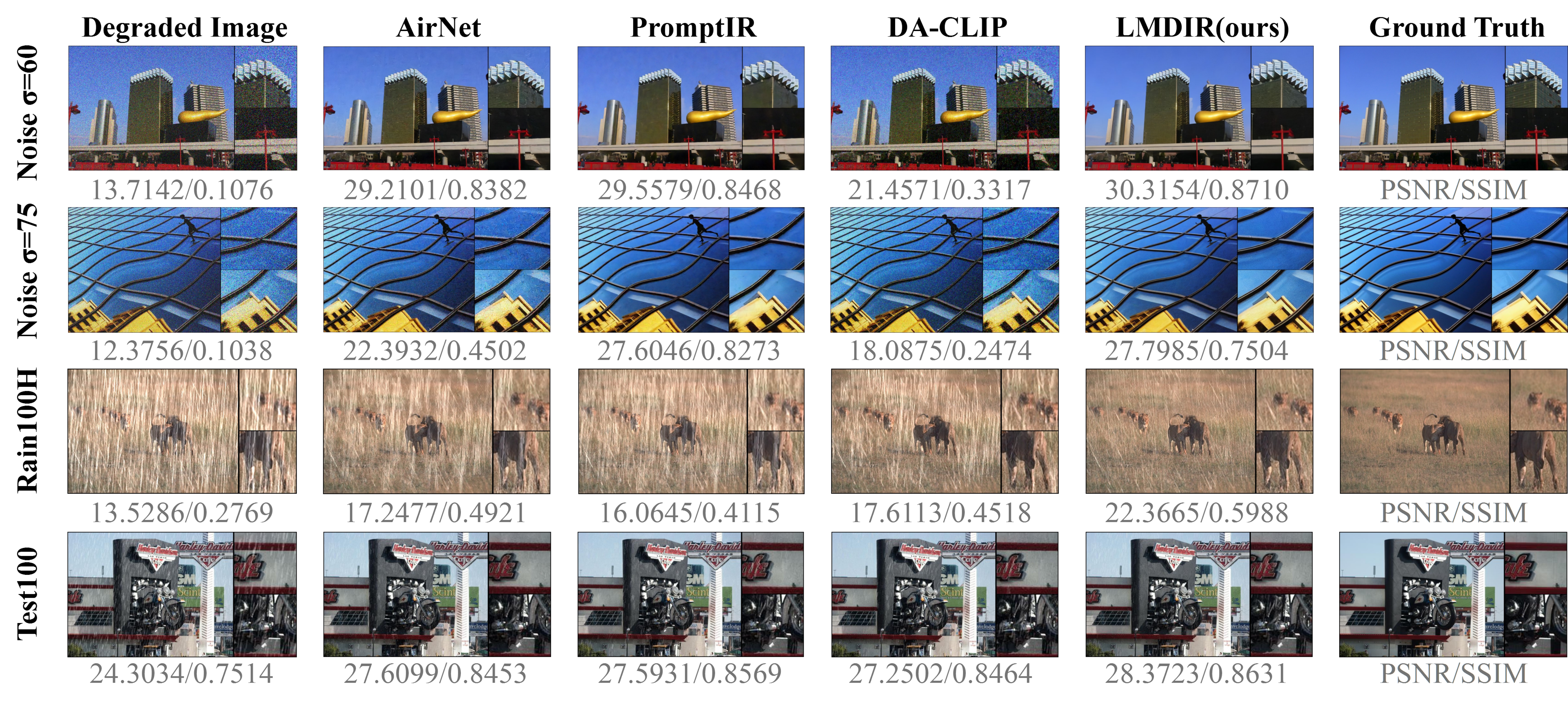}
    \caption{Visual comparison of multiple-in-one methods on OOD dataset.}
    \label{fig:ood_result}
\end{figure*}
 \begin{figure*}[!h]
    \centering
    \includegraphics[width=\linewidth]{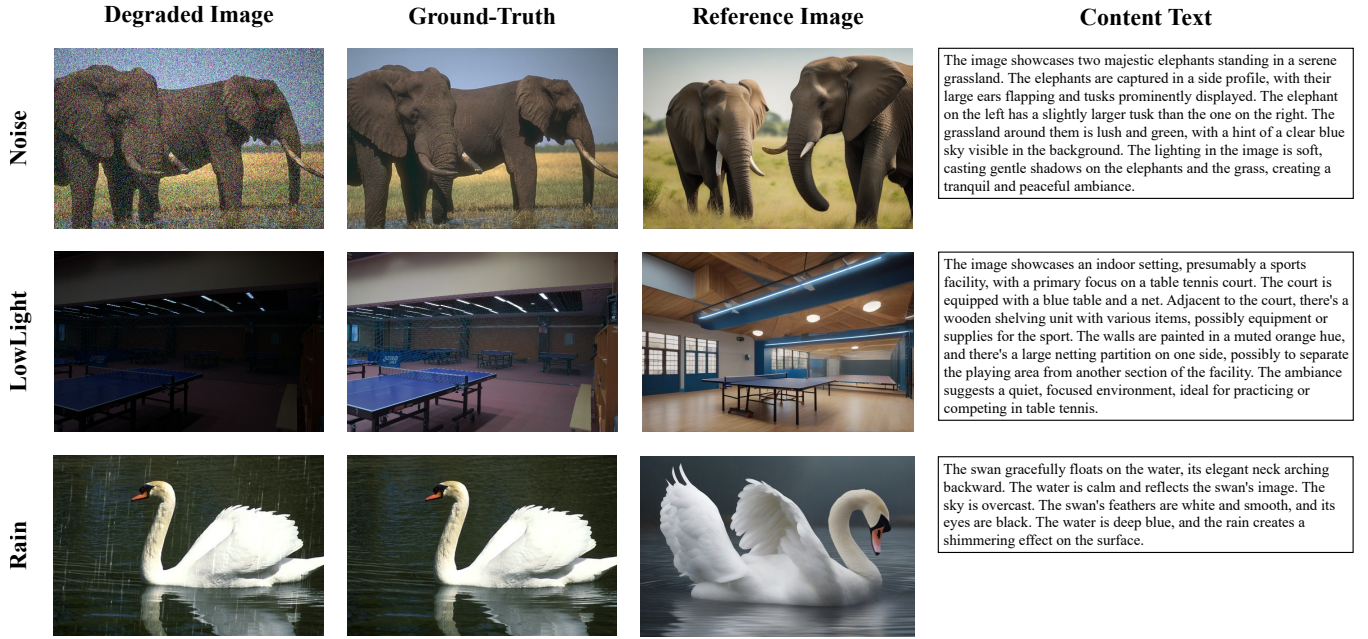}
    \caption{The reference image generated by the diffusion model, given the content text as prompt.}
    \label{fig:reference}
\end{figure*}

\subsubsection{Model generalization performance}
Furthermore, we conducted an additional evaluation of the generalization capabilities of various multi-in-one restoration models on out-of-distribution (OOD) data, thereby evaluating their practical performance in real-world applications. We analyzed the impact of varying noise and rain streak intensities on image restoration tasks. More specifically, for the denoising task, we selected two distinct noise level, 60 and 75. In the context of image deraining, we opted for the Rain100H~\cite{yang2017deep} and Test100~\cite{yang2017deep} datasets as our testing datasets, both of which differ substantially from Rain100L and feature more intense rainfall conditions. The results of these experiments are presented in Table~\ref{table:gen} and Figure~\ref{fig:ood_result}. Our results indicate that the performance of both AirNet and DA-CLIP, whose degradation knowledge is solely based on limited classification knowledge, significantly diminishes when confronted with OOD data. In contrast, the implicit degradation representation of PromptIR exhibits a certain degree of adaptability to OOD data, thereby outperforming AirNet in OOD dataset significantly. Our proposed methods, which combines the prior knowledge of large-scale models with the inherited information present in degraded images, demonstrates an enhanced performance in the presence of OOD data.
\begin{table*}[!h]
\centering
\normalsize
 	\renewcommand{\tabcolsep}{1.5pt} 
\renewcommand{\arraystretch}{1.2}
\caption{Performance on unseen noise level of ($\sigma$ = 60, 75) and severe rain conditions from the Rain100H and test100 dataset. PSNR/SSIM values are reported. The best results are marked in \textbf{bold}.}
\label{table:gen}
\resizebox{\linewidth}{!}{
\begin{tabular}{c|cc|cc|c|c|c}
\hline
\multirow{2}{*}{Method} & \multicolumn{2}{c|}{Denoise(BSD68)} & \multicolumn{2}{c|}{Denoise(Urban100)} & \multirow{2}{*}{Derain(Rain100H)} & \multirow{2}{*}{Derain(Test100)} & \multirow{2}{*}{Average} \\
                        & $\sigma$=60                & $\sigma$=75              & $\sigma$=60                 & $\sigma$=75                &                                   &                                  &                          \\ \hline
AirNet                  & 26.11/0.715       & 20.87/0.421     & 26.38/0.782       & 21.04/0.495       & 15.13/0.508                      & 21.92/0.698                      & 21.91/0.603              \\
PromptIR                & 26.72/0.746       & 23.75/0.569     & 26.57/0.802        & 23.85/0.655    & 13.60/0.416                       & 21.91/0.692                      & 22.73/0.647              \\
DA-CLIP                 & 22.18/0.454      & 19.92/0.301     & 22.21/0.540         & 19.65/0.419       & 16.17/0.509                       & 21.71/0.674                      & 20.31/0.483              \\ \hline
Ours                    & \textbf{27.24}/\textbf{0.761}       & \textbf{24.96}/\textbf{0.625}     & \textbf{27.87}/\textbf{0.825}        & \textbf{25.28}/{\textbf{0.693}}       & \textbf{17.51}/\textbf{0.552}                       & \textbf{22.12}/\textbf{0.701}                      & \textbf{24.16}/\textbf{0.693}              \\ \hline
\end{tabular}}
\end{table*}
\begin{table*}[h]
\centering
\normalsize
 	\renewcommand{\tabcolsep}{2pt} 
\renewcommand{\arraystretch}{1.3}
\caption{Ablation Experiment Results Evaluated with PSNR/SSIM Values. Best results are marked in bold.}
\label{abl}

\resizebox{\linewidth}{!}{
\begin{tabular}{cccc|ccc|ccc|c|c}
\hline
\multicolumn{1}{c|}{}                         &                               &                           &                             & \multicolumn{3}{c|}{Denoise(BSD68)}                  & \multicolumn{3}{c|}{Denoise(Urban100)}                                                               &                          &                                    \\
\multicolumn{1}{c|}{\multirow{-2}{*}{Config}} & \multirow{-2}{*}{degradation} & \multirow{-2}{*}{content} & \multirow{-2}{*}{reference} & $\sigma$=15           & $\sigma$=25          & {$\sigma$=50} & $\sigma$=15                              & $\sigma$=25                              & {$\sigma$=50}          & \multirow{-2}{*}{Derain} & \multirow{-2}{*}{low light}        \\ \hline

\multicolumn{1}{c|}{(I)}                      & \XSolid                              & \XSolid                          & \XSolid                            & 33.67/0.924  & 31.07/0.879 & 27.86/0.782          & 33.46/0.934  & 31.11/0.904  & 27.80/0.837  & 36.55/0.974             & 21.49/0.822 \\
\multicolumn{1}{c|}{(II)}                     & \Checkmark                             & \XSolid                         & \XSolid                           & 33.73/0.927  & 31.45/0.887 & 27.89/0.786             & 33.61/0.938                     & 32.03/0.919                     & 27.90/0.843                      & 36.84/0.986              & 21.94/0.829                        \\
\multicolumn{1}{c|}{(III)}                    & \Checkmark                             & \Checkmark                         & \XSolid                           & 33.84/0.928  & 31.22/0.882 & 28.04/0.792             & 33.77/0.940                     & 31.74/0.915                     & 28.15/0.866                      & 38.06/0.981              & 22.17/0.836                        \\ \hline

\multicolumn{1}{l}{Ours}                      & \Checkmark                            & \Checkmark                         & \Checkmark                           & \textbf{33.99/0.930} & \textbf{31.36/0.885} & \textbf{28.13/0.798}             & \multicolumn{1}{l}{\textbf{34.03/0.944}} & \multicolumn{1}{l}{\textbf{31.84/0.919}} & \multicolumn{1}{l|}{\textbf{28.62/0.873}} & \textbf{38.64/0.983}              & \multicolumn{1}{l}{\textbf{23.34}/\textbf{0.850}}   \\ \hline
\end{tabular}}
\end{table*}
\subsection{Visualization of Reference Images}
We showcase the reference images used by our model, as depicted in Figure~\ref{fig:reference}. The visualization encompasses the degraded image, the ground truth image, the reference image generated by the diffusion model, and the image content generated by the MLLM. As per the illustration, it is evident that the content description generated by MLLM aptly encapsulates the semantic information of the image. Moreover, the reference image generated aligns semantically with the ground truth, thereby providing local details.
\begin{figure*}[!ht]
    \centering
    \includegraphics[width=\linewidth]{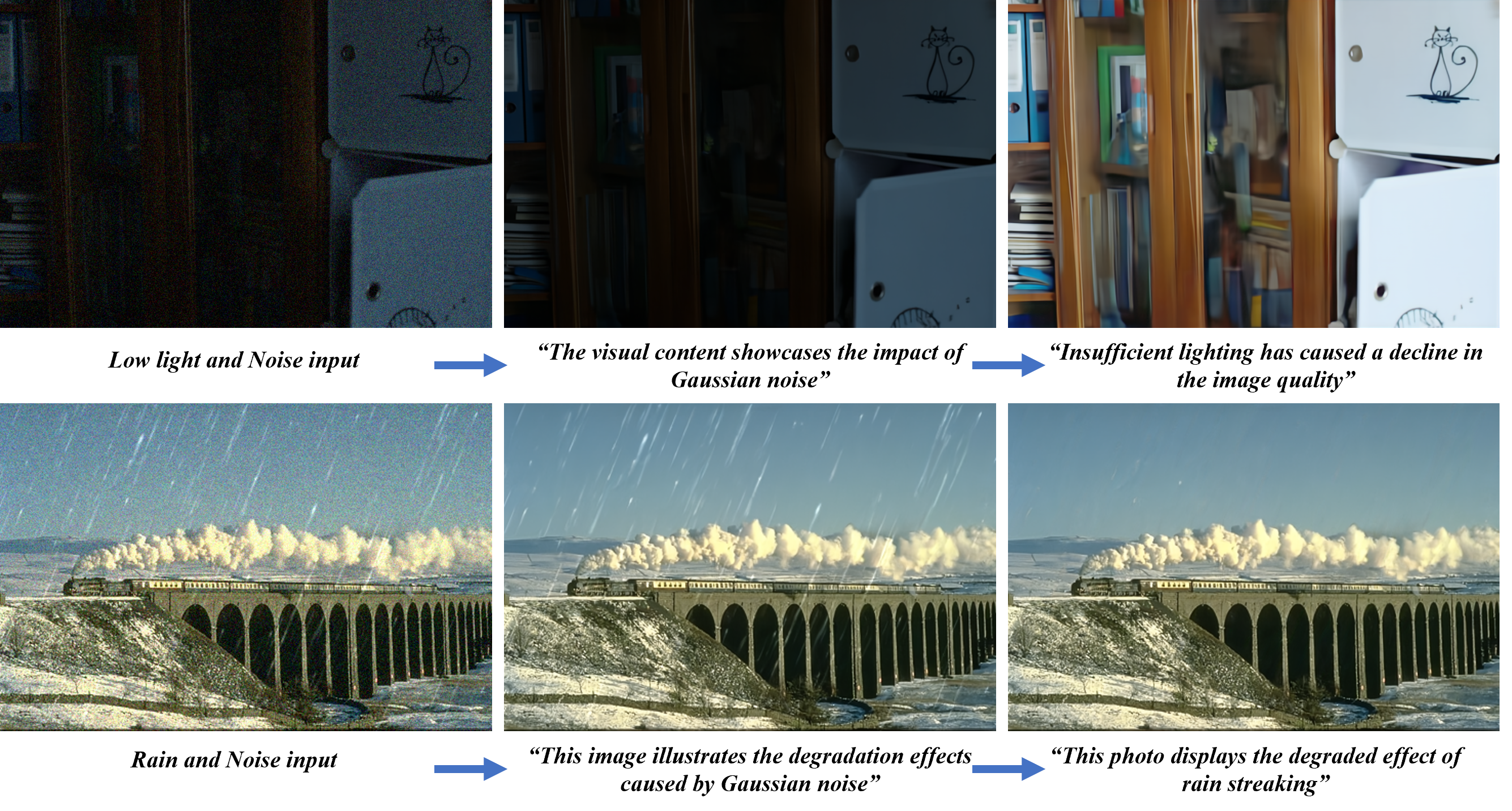}
    \caption{LMDIR can remove particular degradation depending on the human instructions.}
    \label{fig:instruction}
\end{figure*}

\begin{figure*}[!ht]
    \centering
    \includegraphics[width=\linewidth]{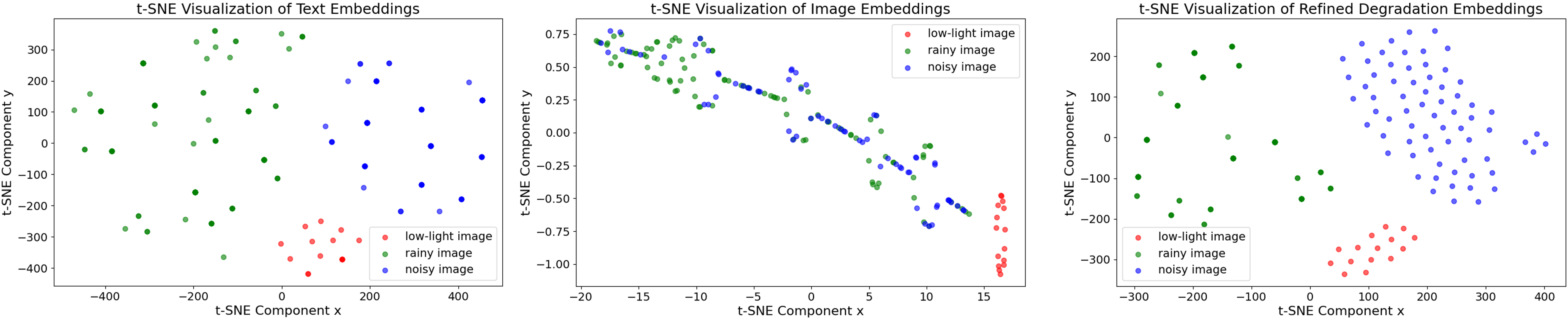}
    \caption{t-SNE visualization of embeddings for different conditions.Left: Text embeddings. Middle: Image embeddings. Right: Refined degradation embeddings provide a clearer separation.}
    \label{fig:tsne}
\end{figure*}
\begin{figure*}[!ht]
    \centering
    \includegraphics[width=\linewidth]{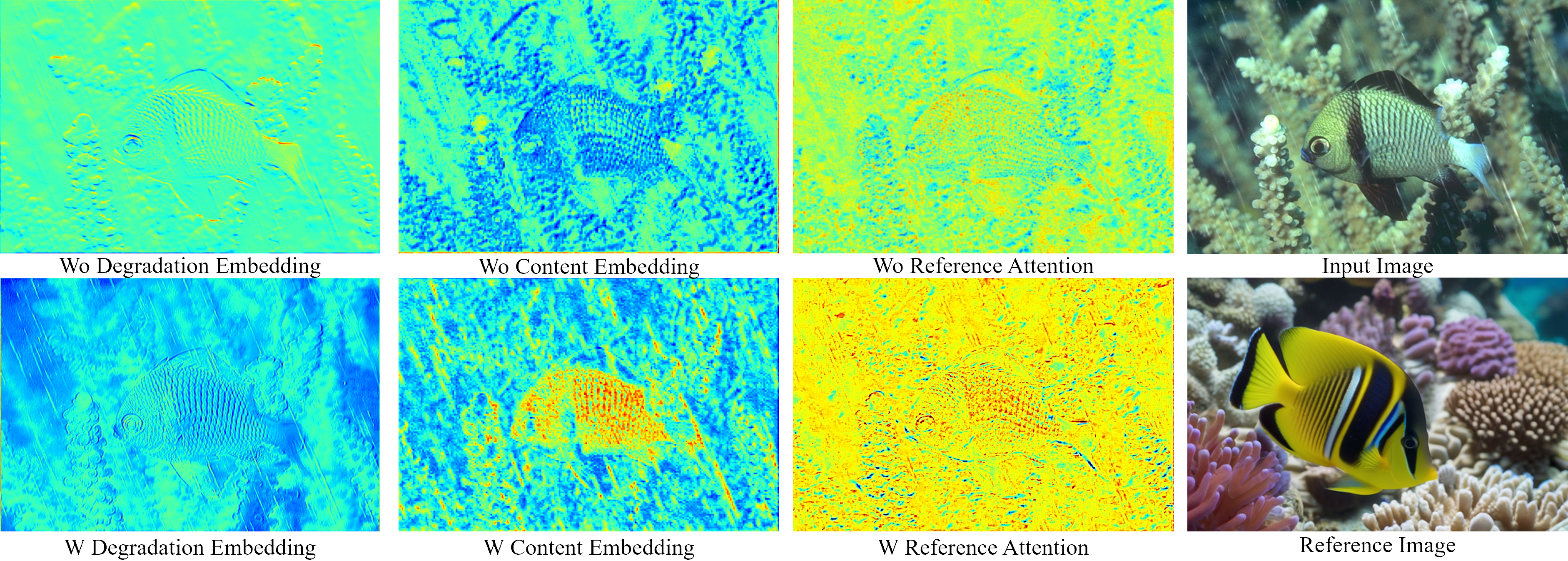}
    \caption{The visualization of feature map of our model. The degradation embedding is observed to concentrate the feature extraction on the rain streaks, while the content embedding directs the focus towards the primary subject of the image. The reference attention enhances the clarity and sharpness of the resulting feature representation.}
    \label{fig:featuremap}
\end{figure*}
\begin{figure}[!ht]
    \centering
    \includegraphics[width=\linewidth]{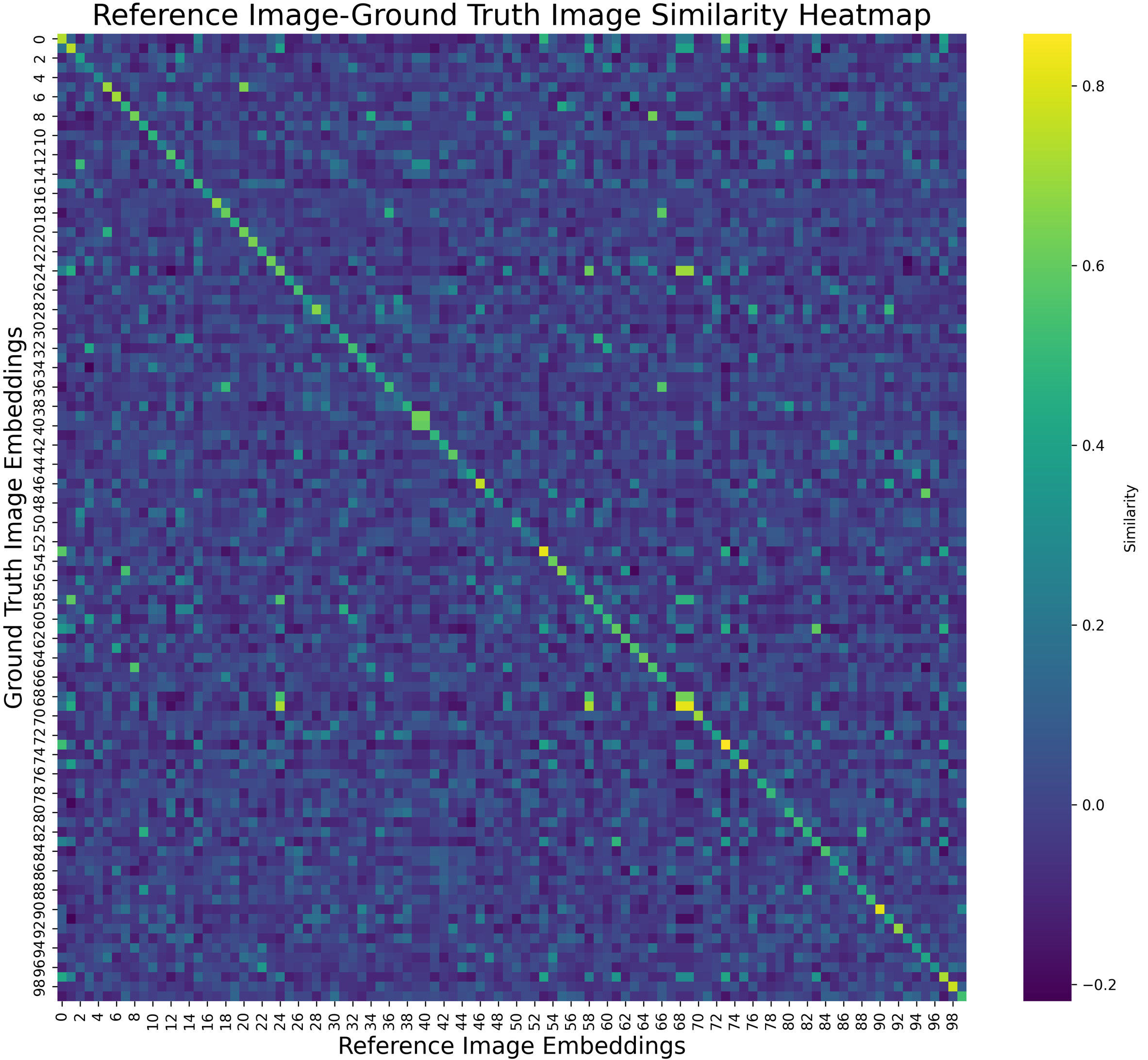}
    \caption{Heatmap illustrating the similarity between reference image embeddings and ground truth image embeddings. The diagonal trend indicates a higher similarity between corresponding images.}
    \label{fig:similar}
\end{figure}

\subsection{On the Effectiveness of User Instructions}
Our proposed model is not only capable of generating degradation prior but also allows for manual user instruction to guide the restoration process. We illustrate this process in Figure~\ref{fig:instruction}. We provided the model with the image affected by mixed degradation. The first row shows the image with low lighting and noise, while the second row shows the image with rain streaks and noise. For these two samples, we manually initially provide a degradation description for noise to obtain a result that removes one type of degradation. Subsequently, we provide a degradation prior of low light or rain to obtain the final reconstructed result. 
\subsection{Ablation Experiment}
To fully evaluate the efficacy of our suggested module, we conducted a series of ablation studies. 
The ablation study was segmented into four distinct parts. Starting from a baseline devoid of any prior, we progressively integrated degradation prior, content prior, and fine-grained image prior. This was done to elucidate the ultimate influence exerted by disparate modules on the overall performance of the model. 
\subsubsection{Baseline}
In this study, we establish a baseline model based on Restormer. This baseline model operates without any reliance on prior information. Leveraging the self-attention inherent capacity for input-adaptive feature extraction, our baseline model demonstrates a Preliminary ability for dealing with diverse degradation, as shown in the first row of Table~\ref{abl}.
\subsubsection{Effectiveness of degradation prior}
Our query-based prompt encoder is designed to enhance the model with a comprehensive degradation prior. In the second set of the ablation study, we
replace the transformer block in encoder part of the baseline model with our degradation-aware transform block, effectively providing the network with degradation-specific information. As evidenced by the second row in the table~\ref{abl}, the introduction of such degradation knowledge led to marked enhancements across multiple task performances, demonstrating the efficacy of the query-based prompt encoder in providing the model with degradation knowledge.
\subsubsection{Effectiveness of textual content prior}
Beyond generating degradation descriptions, our MLLM also produces context description text after receiving visual inputs. These descriptions serve as a source of global prior information, bolstering the model's ability for scene understanding. Building upon Model (I), we replaced the transformer block of bottleneck component within the UNet with our reference-based transformer block. The results presented in the third row of the table reveals that, following the integration of contextual information, the model exhibits marked advancements across a various tasks, thereby affirming the pivotal role of content priors in enhancing overall model performance.

\subsubsection{Effectiveness of local image prior}
Building upon the Model (II), we incorporate fine-grained image priors into our network. We replace the Transformer block in the decoder part of Model(II) with our reference-based Transformer block. We harness a MLLM to generate contextually rich descriptions which serve as prompts for SDXL. This process yields high-fidelity images that retain the identical semantic content as the input image, yet are devoid of any degradation. The integration of such high-quality images plays a pivotal role in offering detailed guidance to the model. As evidenced by the results in the fourth row of the table~\ref{abl}, the integration of fine-grained prior knowledge has led to an improvement in the overall performance of the model.
Furthermore, we conduct a visual analysis of the similarity map between the CLIP features of the ground truth image and the reference image using the Test100 dataset. As illustrated in Fig.~\ref{fig:similar}, the reference image demonstrates a high degree of similarity in the CLIP feature space. This observation provides strong evidence for the efficacy of our generated reference image and the associated content description text.
\subsubsection{Effectiveness of query-based prompt encoder}
Our query-based prompt encoder is motivated by the observation that neither text embeddings nor image embeddings alone can fully capture the information necessary to discriminate between different degradations, as illustrated in Fig.~\ref{fig:tsne}. We employ t-SNE~\cite{van2008visualizing} to visualize the distributions of $e_d$ and $I_d$. The results clearly demonstrate that both $I_d$ and $e_d$ individually fail to effectively distinguish between various degradations. However, after processing through our query-based prompt encoder, the boundaries between degradation types become distinctly delineated. This transformation provides strong evidence for the efficacy of our proposed query-based prompt encoder.
\subsection{Visualization of feature maps}
To underscore the efficacy of our LMDIR approach, we present a visual representation of the feature maps produced by the key components of our proposed architecture in Figure~\ref{fig:featuremap}. This illustration demonstrates the functionality of our designed blocks.
Upon the integration of the global degradation knowledge, the feature map exhibits a pronounced emphasis on the degradation details, highlighting the model's ability to comprehend the image impairments. Incorporating the content information from the scene descriptions further enhances the model's capacity to discern the primary subject within the image.
Moreover, the incorporation of the fine-grained reference image priors imparts a heightened level of sharpness and clarity to the feature representation, demonstrating the complementary benefits of the multi-modal priors utilized in our LMDIR framework.
These visualizations underscore the efficacy of our proposed approach in leveraging the synergistic combination of the MMLMs' generic knowledge and the diffusion models' generative capabilities to enable robust and versatile image restoration, overcoming the limitations of specialized models in dynamic degradation scenarios.

\section{Conclusion}
In this research, we proposed a novel multiple-in-one image restoration framework, termed LMDIR. This approach capitalizes on the wealth of prior knowledge offered by both MLLM and diffusion models. To integrate this information, we carefully tailored a query-based prompt encoder, a reference-based transformer block, a content aware transformer block and a degradation-aware transformer block. Extensive experiments conducted across a diverse range of datasets demonstrate that our proposed method not only surpasses existing state-of-the-art techniques but also exhibits remarkable generalization capabilities on out-of-distribution datasets, showing the superior ability of leveraging large models prior to low-level tasks. 

\bibliographystyle{abbrv}
\bibliography{sample-base}

\begin{thebibliography}{10}

\bibitem{gpt4o}
Hello, gpt-4.
\newblock \url{https://openai.com/index/hello-gpt-4/}, 2024.
\newblock Accessed: 2024-06-24.

\bibitem{2009Variational}
S.~D. Babacan, R.~Molina, and A.~K. Katsaggelos.
\newblock Variational bayesian blind deconvolution using a total variation prior.
\newblock {\em IEEE Transactions on Image Processing}, 18(1):12--26, 2009.

\bibitem{brack2024ledits++}
M.~Brack, F.~Friedrich, K.~Kornmeier, L.~Tsaban, P.~Schramowski, K.~Kersting, and A.~Passos.
\newblock Ledits++: Limitless image editing using text-to-image models.
\newblock In {\em Proceedings of the IEEE/CVF Conference on Computer Vision and Pattern Recognition}, pages 8861--8870, 2024.

\bibitem{brooks2023instructpix2pix}
T.~Brooks, A.~Holynski, and A.~A. Efros.
\newblock Instructpix2pix: Learning to follow image editing instructions.
\newblock In {\em Proceedings of the IEEE/CVF Conference on Computer Vision and Pattern Recognition}, pages 18392--18402, 2023.

\bibitem{chen2023masked}
H.~Chen, J.~Gu, Y.~Liu, S.~A. Magid, C.~Dong, Q.~Wang, H.~Pfister, and L.~Zhu.
\newblock Masked image training for generalizable deep image denoising.
\newblock In {\em Proceedings of the IEEE/CVF Conference on Computer Vision and Pattern Recognition}, pages 1692--1703, 2023.

\bibitem{chen2021pre}
H.~Chen, Y.~Wang, T.~Guo, C.~Xu, Y.~Deng, Z.~Liu, S.~Ma, C.~Xu, C.~Xu, and W.~Gao.
\newblock Pre-trained image processing transformer.
\newblock In {\em Proceedings of the IEEE/CVF conference on computer vision and pattern recognition}, pages 12299--12310, 2021.

\bibitem{chen2022simple}
L.~Chen, X.~Chu, X.~Zhang, and J.~Sun.
\newblock Simple baselines for image restoration.
\newblock In {\em European conference on computer vision}, pages 17--33. Springer, 2022.

\bibitem{chen2021hinet}
L.~Chen, X.~Lu, J.~Zhang, X.~Chu, and C.~Chen.
\newblock Hinet: Half instance normalization network for image restoration.
\newblock In {\em Proceedings of the IEEE/CVF Conference on Computer Vision and Pattern Recognition}, pages 182--192, 2021.

\bibitem{desnowing}
W.-T. Chen, H.-Y. Fang, C.-L. Hsieh, C.-C. Tsai, I.~Chen, J.-J. Ding, S.-Y. Kuo, et~al.
\newblock All snow removed: Single image desnowing algorithm using hierarchical dual-tree complex wavelet representation and contradict channel loss.
\newblock In {\em Proceedings of the IEEE/CVF International Conference on Computer Vision}, pages 4196--4205, 2021.

\bibitem{crowson2022vqgan}
K.~Crowson, S.~Biderman, D.~Kornis, D.~Stander, E.~Hallahan, L.~Castricato, and E.~Raff.
\newblock Vqgan-clip: Open domain image generation and editing with natural language guidance.
\newblock In {\em European Conference on Computer Vision}, pages 88--105. Springer, 2022.

\bibitem{geva2020transformer}
M.~Geva, R.~Schuster, J.~Berant, and O.~Levy.
\newblock Transformer feed-forward layers are key-value memories.
\newblock {\em arXiv preprint arXiv:2012.14913}, 2020.

\bibitem{zerodce}
C.~Guo, C.~Li, J.~Guo, C.~C. Loy, J.~Hou, S.~Kwong, and R.~Cong.
\newblock Zero-reference deep curve estimation for low-light image enhancement.
\newblock In {\em Proceedings of the IEEE/CVF conference on computer vision and pattern recognition}, pages 1780--1789, 2020.

\bibitem{huang2015single}
J.-B. Huang, A.~Singh, and N.~Ahuja.
\newblock Single image super-resolution from transformed self-exemplars.
\newblock In {\em Proceedings of the IEEE conference on computer vision and pattern recognition}, pages 5197--5206, 2015.

\bibitem{rain2}
K.~Jiang, Z.~Wang, P.~Yi, C.~Chen, B.~Huang, Y.~Luo, J.~Ma, and J.~Jiang.
\newblock Multi-scale progressive fusion network for single image deraining.
\newblock In {\em Proceedings of the IEEE/CVF conference on computer vision and pattern recognition}, pages 8346--8355, 2020.

\bibitem{jiang2021robust}
Y.~Jiang, K.~C. Chan, X.~Wang, C.~C. Loy, and Z.~Liu.
\newblock Robust reference-based super-resolution via c2-matching.
\newblock In {\em Proceedings of the IEEE/CVF Conference on Computer Vision and Pattern Recognition}, pages 2103--2112, 2021.

\bibitem{kawar2023imagic}
B.~Kawar, S.~Zada, O.~Lang, O.~Tov, H.~Chang, T.~Dekel, I.~Mosseri, and M.~Irani.
\newblock Imagic: Text-based real image editing with diffusion models.
\newblock In {\em Proceedings of the IEEE/CVF Conference on Computer Vision and Pattern Recognition}, pages 6007--6017, 2023.

\bibitem{kong2024towards}
X.~Kong, C.~Dong, and L.~Zhang.
\newblock Towards effective multiple-in-one image restoration: A sequential and prompt learning strategy.
\newblock {\em arXiv preprint arXiv:2401.03379}, 2024.

\bibitem{lee2021brief}
H.~Lee, U.~Ullah, J.-S. Lee, B.~Jeong, and H.-C. Choi.
\newblock A brief survey of text driven image generation and maniulation.
\newblock In {\em 2021 IEEE International Conference on Consumer Electronics-Asia (ICCE-Asia)}, pages 1--4. IEEE, 2021.

\bibitem{li2022all}
B.~Li, X.~Liu, P.~Hu, Z.~Wu, J.~Lv, and X.~Peng.
\newblock All-in-one image restoration for unknown corruption.
\newblock In {\em Proceedings of the IEEE/CVF Conference on Computer Vision and Pattern Recognition}, pages 17452--17462, 2022.

\bibitem{li2020all}
R.~Li, R.~T. Tan, and L.-F. Cheong.
\newblock All in one bad weather removal using architectural search.
\newblock In {\em Proceedings of the IEEE/CVF conference on computer vision and pattern recognition}, pages 3175--3185, 2020.

\bibitem{li2023ntire}
Y.~Li, Y.~Zhang, R.~Timofte, L.~Van~Gool, Z.~Tu, K.~Du, H.~Wang, H.~Chen, W.~Li, X.~Wang, et~al.
\newblock Ntire 2023 challenge on image denoising: Methods and results.
\newblock In {\em Proceedings of the IEEE/CVF Conference on Computer Vision and Pattern Recognition}, pages 1904--1920, 2023.

\bibitem{liang2021swinir}
J.~Liang, J.~Cao, G.~Sun, K.~Zhang, L.~Van~Gool, and R.~Timofte.
\newblock Swinir: Image restoration using swin transformer.
\newblock In {\em Proceedings of the IEEE/CVF international conference on computer vision}, pages 1833--1844, 2021.

\bibitem{luo2023controlling}
Z.~Luo, F.~K. Gustafsson, Z.~Zhao, J.~Sj{\"o}lund, and T.~B. Sch{\"o}n.
\newblock Controlling vision-language models for universal image restoration.
\newblock {\em arXiv preprint arXiv:2310.01018}, 2023.

\bibitem{ma2023prores}
J.~Ma, T.~Cheng, G.~Wang, Q.~Zhang, X.~Wang, and L.~Zhang.
\newblock Prores: Exploring degradation-aware visual prompt for universal image restoration.
\newblock {\em arXiv preprint arXiv:2306.13653}, 2023.

\bibitem{wed}
D.~Martin, C.~Fowlkes, D.~Tal, and J.~Malik.
\newblock A database of human segmented natural images and its application to evaluating segmentation algorithms and measuring ecological statistics.
\newblock In {\em Proceedings Eighth IEEE International Conference on Computer Vision. ICCV 2001}, volume~2, pages 416--423. IEEE, 2001.

\bibitem{bsd68}
D.~Martin, C.~Fowlkes, D.~Tal, and J.~Malik.
\newblock A database of human segmented natural images and its application to evaluating segmentation algorithms and measuring ecological statistics.
\newblock In {\em Proc. 8th Int'l Conf. Computer Vision}, volume~2, pages 416--423, July 2001.

\bibitem{podell2023sdxl}
D.~Podell, Z.~English, K.~Lacey, A.~Blattmann, T.~Dockhorn, J.~M{\"u}ller, J.~Penna, and R.~Rombach.
\newblock Sdxl: Improving latent diffusion models for high-resolution image synthesis.
\newblock {\em arXiv preprint arXiv:2307.01952}, 2023.

\bibitem{potlapalli2023promptir}
V.~Potlapalli, S.~W. Zamir, S.~Khan, and F.~S. Khan.
\newblock Promptir: Prompting for all-in-one blind image restoration.
\newblock {\em arXiv preprint arXiv:2306.13090}, 2023.

\bibitem{clip}
A.~Radford, J.~W. Kim, C.~Hallacy, A.~Ramesh, G.~Goh, S.~Agarwal, G.~Sastry, A.~Askell, P.~Mishkin, J.~Clark, et~al.
\newblock Learning transferable visual models from natural language supervision.
\newblock In {\em International conference on machine learning}, pages 8748--8763. PMLR, 2021.

\bibitem{rain1}
D.~Ren, W.~Zuo, Q.~Hu, P.~Zhu, and D.~Meng.
\newblock Progressive image deraining networks: A better and simpler baseline.
\newblock In {\em Proceedings of the IEEE/CVF conference on computer vision and pattern recognition}, pages 3937--3946, 2019.

\bibitem{7473901}
W.~Ren, X.~Cao, J.~Pan, X.~Guo, W.~Zuo, and M.-H. Yang.
\newblock Image deblurring via enhanced low-rank prior.
\newblock {\em IEEE Transactions on Image Processing}, 25(7):3426--3437, 2016.

\bibitem{ren2019low}
W.~Ren, S.~Liu, L.~Ma, Q.~Xu, X.~Xu, X.~Cao, J.~Du, and M.-H. Yang.
\newblock Low-light image enhancement via a deep hybrid network.
\newblock {\em IEEE Transactions on Image Processing}, 28(9):4364--4375, 2019.

\bibitem{rombach2022high}
R.~Rombach, A.~Blattmann, D.~Lorenz, P.~Esser, and B.~Ommer.
\newblock High-resolution image synthesis with latent diffusion models.
\newblock In {\em Proceedings of the IEEE/CVF conference on computer vision and pattern recognition}, pages 10684--10695, 2022.

\bibitem{sheynin2024emu}
S.~Sheynin, A.~Polyak, U.~Singer, Y.~Kirstain, A.~Zohar, O.~Ashual, D.~Parikh, and Y.~Taigman.
\newblock Emu edit: Precise image editing via recognition and generation tasks.
\newblock In {\em Proceedings of the IEEE/CVF Conference on Computer Vision and Pattern Recognition}, pages 8871--8879, 2024.

\bibitem{deblurring}
X.~Tao, H.~Gao, X.~Shen, J.~Wang, and J.~Jia.
\newblock Scale-recurrent network for deep image deblurring.
\newblock In {\em Proceedings of the IEEE conference on computer vision and pattern recognition}, pages 8174--8182, 2018.

\bibitem{van2008visualizing}
L.~Van~der Maaten and G.~Hinton.
\newblock Visualizing data using t-sne.
\newblock {\em Journal of machine learning research}, 9(11), 2008.

\bibitem{wei1808deep}
C.~Wei, W.~Wang, W.~Yang, and J.~Liu.
\newblock Deep retinex decomposition for low-light enhancement. arxiv 2018.
\newblock {\em arXiv preprint arXiv:1808.04560}, 1808.

\bibitem{wu2023q}
H.~Wu, Z.~Zhang, E.~Zhang, C.~Chen, L.~Liao, A.~Wang, C.~Li, W.~Sun, Q.~Yan, G.~Zhai, et~al.
\newblock Q-bench: A benchmark for general-purpose foundation models on low-level vision.
\newblock {\em arXiv preprint arXiv:2309.14181}, 2023.

\bibitem{yang2020learning}
F.~Yang, H.~Yang, J.~Fu, H.~Lu, and B.~Guo.
\newblock Learning texture transformer network for image super-resolution.
\newblock In {\em Proceedings of the IEEE/CVF conference on computer vision and pattern recognition}, pages 5791--5800, 2020.

\bibitem{rain1800}
W.~Yang, R.~T. Tan, J.~Feng, Z.~Guo, S.~Yan, and J.~Liu.
\newblock Joint rain detection and removal from a single image with contextualized deep networks.
\newblock {\em IEEE transactions on pattern analysis and machine intelligence}, 42(6):1377--1393, 2019.

\bibitem{yang2017deep}
W.~Yang, R.~T. Tan, J.~Feng, J.~Liu, Z.~Guo, and S.~Yan.
\newblock Deep joint rain detection and removal from a single image.
\newblock In {\em Proceedings of the IEEE conference on computer vision and pattern recognition}, pages 1357--1366, 2017.

\bibitem{yin2023survey}
S.~Yin, C.~Fu, S.~Zhao, K.~Li, X.~Sun, T.~Xu, and E.~Chen.
\newblock A survey on multimodal large language models.
\newblock {\em arXiv preprint arXiv:2306.13549}, 2023.

\bibitem{zamir2022restormer}
S.~W. Zamir, A.~Arora, S.~Khan, M.~Hayat, F.~S. Khan, and M.-H. Yang.
\newblock Restormer: Efficient transformer for high-resolution image restoration.
\newblock In {\em Proceedings of the IEEE/CVF conference on computer vision and pattern recognition}, pages 5728--5739, 2022.

\bibitem{zhang2019image}
Z.~Zhang, Z.~Wang, Z.~Lin, and H.~Qi.
\newblock Image super-resolution by neural texture transfer.
\newblock In {\em Proceedings of the IEEE/CVF conference on computer vision and pattern recognition}, pages 7982--7991, 2019.

\bibitem{zheng2018crossnet}
H.~Zheng, M.~Ji, H.~Wang, Y.~Liu, and L.~Fang.
\newblock Crossnet: An end-to-end reference-based super resolution network using cross-scale warping.
\newblock In {\em Proceedings of the European conference on computer vision (ECCV)}, pages 88--104, 2018.

\bibitem{zhou2022conditional}
K.~Zhou, J.~Yang, C.~C. Loy, and Z.~Liu.
\newblock Conditional prompt learning for vision-language models.
\newblock In {\em Proceedings of the IEEE/CVF conference on computer vision and pattern recognition}, pages 16816--16825, 2022.

\end{thebibliography}

\vfill

\end{document}